# MatES: Web-based Forward Chaining Expert System for Maternal Care

*Haile Misgna[1], Moges Ahmed[2], Anubhav Kumar[3]*

[1,2,3]School of Computing, EiT-M, Mekelle University, Ethiopia

Corresponding author: [1]haile.misgna@mu.edu.et

**Abstract**

The solution to prevent maternal complications are known and preventable by trained health professionals. But in countries like Ethiopia where the patient to physician ratio is 1 doctor to 1000 patients, maternal mortality and morbidity rate is high. To fill the gap of highly trained health professionals, Ethiopia introduced health extension programs. Task shifting to health extension workers (HEWs) contributed in decreasing mortality and morbidity rate in Ethiopia. Knowledge-gap has been one of the major challenges to HEWs. The reasons are trainings are not given in regular manner, there is no midwife, gynecologists or doctors around for consultation, and all guidelines are paper-based which are easily exposed to damage. In this paper, we describe the design and implementation of a web-based expert system for maternal care. We only targeted the major 10 diseases and complication of maternal health issues seen in Sub-Saharan Africa. The expert system can be accessed through the use of web browsers from computers as well as smart phones. Forward chaining rule-based expert system is used in order to give suggestions and create a new knowledge from the knowledge-base. This expert system can be used to train HEWs in the field of maternal health.

1. **Introduction**

Maternal health is about the health of women starting from preconception up to the postpartum period. Its aim is to give positive experience of motherhood by reducing morbidity and mortality rate of mothers and infants. But attaining this positive experience has been challenged by maternal health complications. Every day 810 women are dying because of preventable pregnancy and childbirth complications, and the 99% of the death reports are from developing countries (WHO, 2019). The major maternal mortality complications are well-known; severe bleeding, infection and pre-eclampsia (WHO, 2019). In Ethiopia, most of the causes for maternal mortality are postpartum haemorrhage, sepsis, eclampsia, obstructed labor and unsafe abortion (Desta, 2017). Majority of the causes for maternal mortality are preventable.

Lack of enough number of health professionals, weak infrastructure and poor supply chain played a big role in challenging the health sector in developing countries (Bilal, 2011). In Ethiopia, the Federal Ministry of Health (FMOH) introduced task shifting program in order to reduce the impact of lack of trained health professionals. The introduction of health extension workers (HEWs) to the health sector filled the gap seen in the number of trained health professionals. The HEWs are engaged in treating specific tasks with short training and fewer qualifications. They are recruited from once community and trained on health education and communication, hygiene and environmental sanitation, disease prevention and control, and family health (Desta, 2017). With the help of HEWs, Ethiopia decreased child and maternal mortality rate by large percent in the last 30 years (Berhan, 2014).

Knowledge-gap has been one of the major challenges experienced by the HEWs. This is raised because of the lack of regular trainings, no highly trained health professional around for consultation, and short life span of paper-based maternal health guidelines. Most of the time, trainings are given one time after the recruitment. They are not periodical. This one-time training is important for the HEWs but giving trainings regularly will help the workers to be at the top of their game. The other challenge is HEWs are deployed in the rural part of Ethiopia, where there is no a hospital or a clinic nearby. The possibility to consult a doctor, gynecologists or midwife is very rare. The workers are highly dependent on paper-based maternal health guidelines which is easily exposed to a damage.

In this work, we propose MatES, a web-based forward chain rule-based expert system that can be installed on smartphones or accessed using smartphones through the use of Internet. This will address the knowledge-gap challenge by providing a service of consultancy and by being a training platform for HEWs. Section 2 of this document discusses about expert systems. In section 3, the development and design of MatES is thoroughly discussed. Section 4 discusses the implementation and presents snapshots of the system. Our recommendation and conclusion is discussed on section 5.

## 2. Expert System

Expert system is an application of artificial intelligence that simulates the decision making ability of human experts. It is a computer software that has 3 major components; user interface, inference engine and the knowledge-base. The user interface allows a smooth communication between the end user and the expert system. This component of the system has to have easy, simple, and not confusing user interface. The inference engine is another component of an expert system that is responsible for reasoning and new knowledge creation. The inference engine fires set of actions based on collected inputs from a given environment. It has the capability to infer a new knowledge from previously known knowledge resided in the knowledge-base (KB). The knowledge-base is a resident for the set of facts, rules and objects. The KB stores the expertise of human expert in the form of rules and facts engineered by knowledge engineers. To map inputs collected from the user interface with knowledge in the knowledge-base and to form a new knowledge from the knowledge-base, the inference engine relates.

### 2.1. Rule-based Expert System

Rule-based expert systems use rules as a knowledge representation technique. The rules are presented in the form *IF-THEN* statements. The *IF* part is called premise and the *THEN* part is called conclusion. A rule-based system has facts, rules and termination criteria elements. The data and associated conditions are the fact elements. Facts interact with data directly to determine if the event is of interest. The rule element of the expert systems relates facts with actions. In other words, it constructs an *IF-THEN* rule by putting the facts under the *IF* part and the set of actions under the *THEN* part. Complex rules can be formed by joining rules using logical operators. When there are multiple set of facts to be checked individually, *AND*, and *OR* operators are used to form premise part of the rule. A rule can also trigger multiple set of actions. These set of actions can also be joined by logical operators.

### 2.2. Forward Chaining Systems

Forward chaining systems are data-driven rule-based systems that trigger actions based on the facts under the premise part of the rule. They start from the known data and add a new fact to the knowledge-base if it is not already in the knowledge-base. The disadvantage with forward chaining is many rules can be executed even they do have nothing to do with the established goal. So it is not efficient if one fact is only to be inferred. Forward chaining systems perform well when the goal is not known. They can trigger sounding actions if adequate information is gathered.

### 2.3. Web-based Expert Systems

Web-based applications give the advantage of accessibility of systems from different platforms. Once the application is deployed in a server, the system can be accessed through the use of any web browser from any operating system. Web-based expert systems are used in diagnosis and treatment of human (Persulessy, Pratama, Setiawan, & Sevani, 2019) (Ahmed A. Al-Hajji, 2019), animal (Fu Zetian, 2005) (Daoliang Li, 2002) and plant diseases (Fahad Shahbaz Khan, 2008) (Ravisankar, Sivaraju, D, & Rao, 2018).

## 3. Maternal Care Expert System Design and Development

We used a rule-based expert system for maternal health care for the design and implementation of maternal care expert system (MatES). Range of medical related expert systems use rule-based expert systems (FAZEL, 2010) (Seto, 2012). Even though it has been shown that a good experimental result, the systems have been in practice in teaching and training of health professional only. In this work also, we are recommending HEWs to use the system as a training and practice platform.

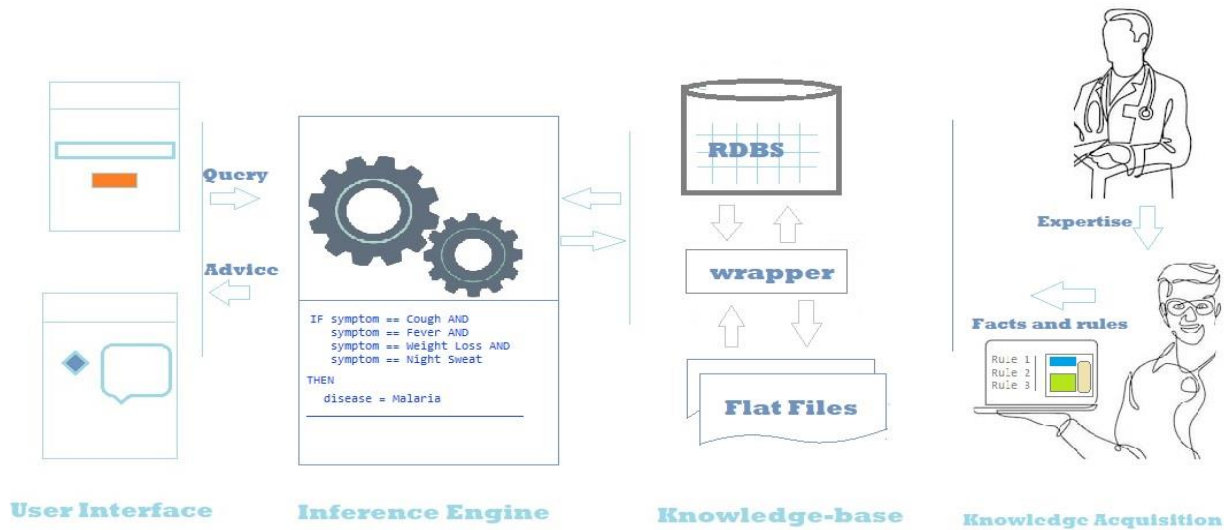

*Figure 1: Structure of expert system*

The proposed expert systems reasons based on symptoms, and diseases. The inference engine of the expert system takes two types of inputs and generates three types of results. The first type of input is symptoms. Taking symtpoms as an input, the expert systems trigers the inference engine to fire a type of diseases. This part of the expert system helps the physician to identfy the type of disease based on the syptoms and some labratories results. Once the diseases is identified, it will be send to the system as the second form of input. This input makes the expert system to fire two different results adjacently. The first result is care and treatment for the disease. This part shows the different levels of a given disease and a corresponding treatement for evey level and stage. And the second fired result is information regarding the consequence of the disease if it is untreated.

$$\frac{\text{Premise}}{\text{Conclusion}} \quad \text{(Rule 1)}$$

The *Rule 1* is read as *if premise then conclusion*. A number of conditions can be included in the premise part of the expresssion using *AND* or/and *OR* connectives based on the compound nature of the input. And the conclusion part of the expression shows the result if the sets in the permise are satified.

MatES considers only 10 major diseases common in sub-Saharan Africa that creates complications during pregnancy. According to the suggestion of CDC and WHO, the following major diseases create pregnancy complications in developing countries (CDC, 2018); AIDS, malaria, TB, Hepatitis B, Hepatitis C, STI, anemia, UTI, mental health conditions, and hypertension. These diseases have more the 40 symptoms in total as it is presented in table 1.

*Table 1: Symptoms list*

| ID | Symptoms | ID | Symptoms | ID | Symptoms |
|---|---|---|---|---|---|
| 1 | Cough | 16 | Pelvic pain | 31 | Urgency to use bathroom often |
| 2 | Fever | 17 | Changes in appetite | 32 | Loss of interest in fun activities |
| 3 | Weight loss | 18 | Concentration problems | 33 | Pain while using bathroom |
| 4 | Night sweat | 19 | Low mood | 34 | Lack of appropriate weight gain |
| 5 | Hemoptysis | 20 | Genital ulcers | 35 | Decision making problems |
| 6 | Malaise | 21 | Abdominal pain | 36 | Feelings of worthlessness |
| 7 | Headache | 22 | Concentrated urine | 37 | Feelings of shame |

| 8  | Myalgia           | 23 | Sad mood                 | 38 | Feelings of guilt             |
|----|-------------------|----|--------------------------|----|-------------------------------|
| 9  | Vomiting          | 24 | Change in sleep          | 39 | Thoughts of not worth living  |
| 10 | Nausea            | 25 | Change in energy         | 40 | Back pain                     |
| 11 | Jaundice          | 26 | Thinking problems        |    |                               |
| 12 | Fatigue           | 27 | Bad smell urine          |    |                               |
| 13 | Shakiness         | 28 | Urine looks cloudy       |    |                               |
| 14 | Feel weak         | 29 | Urine looks reddish      |    |                               |
| 15 | Vaginal discharge | 30 | Pressure in lower belly  |    |                               |

For this work, we developed two rule-based algorithms. As it is shown in algorithm 1, the first algorithm takes a disease and returns care and treatment. The consideration of using disease as input parameter is sometimes the disease may be known by the health worker. And this will help the user of the system to give immediate suggestion of care and treatment. Besides this, the system also provides the consequence of the disease if it is not treated. We add this component to show the health worker and the pregnant woman how serious the complication can be.

Algorithm 1

```
persitent: DN, disease name
          careTreatment, care and treatment for the corresponding disease
          notTreated, what will happen if the disease is not treated
IF DN is given
THEN
     retrieve careTreatment And
     retrieve notTreated
END
```

Algorithm 1 takes disease as premises and fires the care and treatment, and the consequence if it is not treated. The values of *careTreatment*, and *notTreated* are retrieved from the knowledge-base that contains the mapping of diseases, and care and treatment in the form of rules.

$$\frac{Diseases}{Care\ and\ Treatment} \qquad (Rule\ 2)$$

In short, algorithm 1 and rule 2 represents the firing of care and treatment if disease is known. Here is a simple scenario:

**IF** diseases = {TB}
**THEN** care_treatment = {{tablet = {ethambutol AND isoniazid AND rifampcin AND pyrazinamide}, time = {for 2 months}} AND {tablet = isoniazid AND rifampcin}, time = {for 4 months}}}

The above scenario shows that a rule fires care and treatment based on a type of disease. If the disease is TB, then expert system suggests care and treatment for it. In addition, it suggests medication and time duration.

$$\frac{Diseases}{Results\ if\ untreated} \qquad (Rule\ 3)$$

Rule 3 is another form of rule; it shows that a premise with known disease and then the expert system fires consequences of the disease if it is not treated timely. This type of rule shows that what can possibly happen if the disease is untreated timely.

**IF** disease = {TB}
**THEN** result_if_untreated = {delivering a premature baby, low birth weight}

The above scenario shows what will result if a pregnant woman with TB is untreated. TB can result in delivering a premature baby and low birth weight if it is not treated timely.

The other type of input is list of symptoms. When symptoms are given as an input, the system suggests list of diseases that are characterized by the list of selected symptoms. The disease with most list of symptoms selected will be at the top of the suggestions list. Since most of the diseases share similar symptoms, the system suggests list of disease with symptoms that matches with symptoms from the query. It allows the intervention of human experts to look into the list of recommendation and pick one of the suggested diseases or give medication for all or some of the diseases.

$$\frac{Symptoms}{Diseases} \qquad (Rule\ 4)$$

Rule 4 shows list of symptoms as input and a disease as a conclusion. By considering the different set of symptoms, the system fires a corresponding a list of diseases with its corresponding probability.

> *IF symptom = {cough AND weight loss AND night sweats AND fever}*
> *THEN Disease= {TB}*

The above is a scenario that fires a disease type based on the premises. If the mother is showing symptoms like cough, weight loss, night sweat and fever the expert system suggests the diseases is TB.

For the second algorithm, we need to define some formulation before discussing the algorithm in the form of flowchart.

***Definition 1***: A set of *l* number of queries, *Q*, posed by the user, $\{q_1, q_2, q_3, ..., q_l\}$, are represented as follows:

$$\{q_1, q_2, q_3, ..., q_l\} \subseteq Q$$

***Definition 2***: A set of *m* number of symptoms, *S*, of a disease are represented as follows:

$$\{s_1, s_2, s_3, ..., s_m\} \subseteq S$$

***Definition 3***: A set of *n* number of diseases, *D*, are represented as follows:

$$\{d_1, d_2, d_3, ..., d_n\} \subseteq D$$

Diseases and symptoms are presented in the form of matrix, $|D \times S|$. Each entry of the matrix is either 0 or 1.

$$|d_i \times s_j| = \begin{cases} 1, \text{ if } d_i \text{ has } s_j \\ 0, \text{ otherwise} \end{cases}$$

Entry is 1 if $D_i$ has a symptom $S_j$ otherwise the entry is 0. To suggest a disease based on query, Q, measured the probability of each disease based the query posed symptoms. The computation of $P(d_i)$ is the sum of the entries of $|d_i \times s_j|$ if $s_j$ is in the set of posed queries, *Q*.

$$P(d_i) = \sum |d_i \times s_j|, \text{ if } s_j \text{ is in } Q$$

The system recommends the disease with the highest probability. This means the symptom with the most number of symptom queries is suggested.

$$argmax\ P(d_i)$$

Algorithm 2 uses the above concepts to provide a suggestion based on queries collected from the user interface component of the expert system.

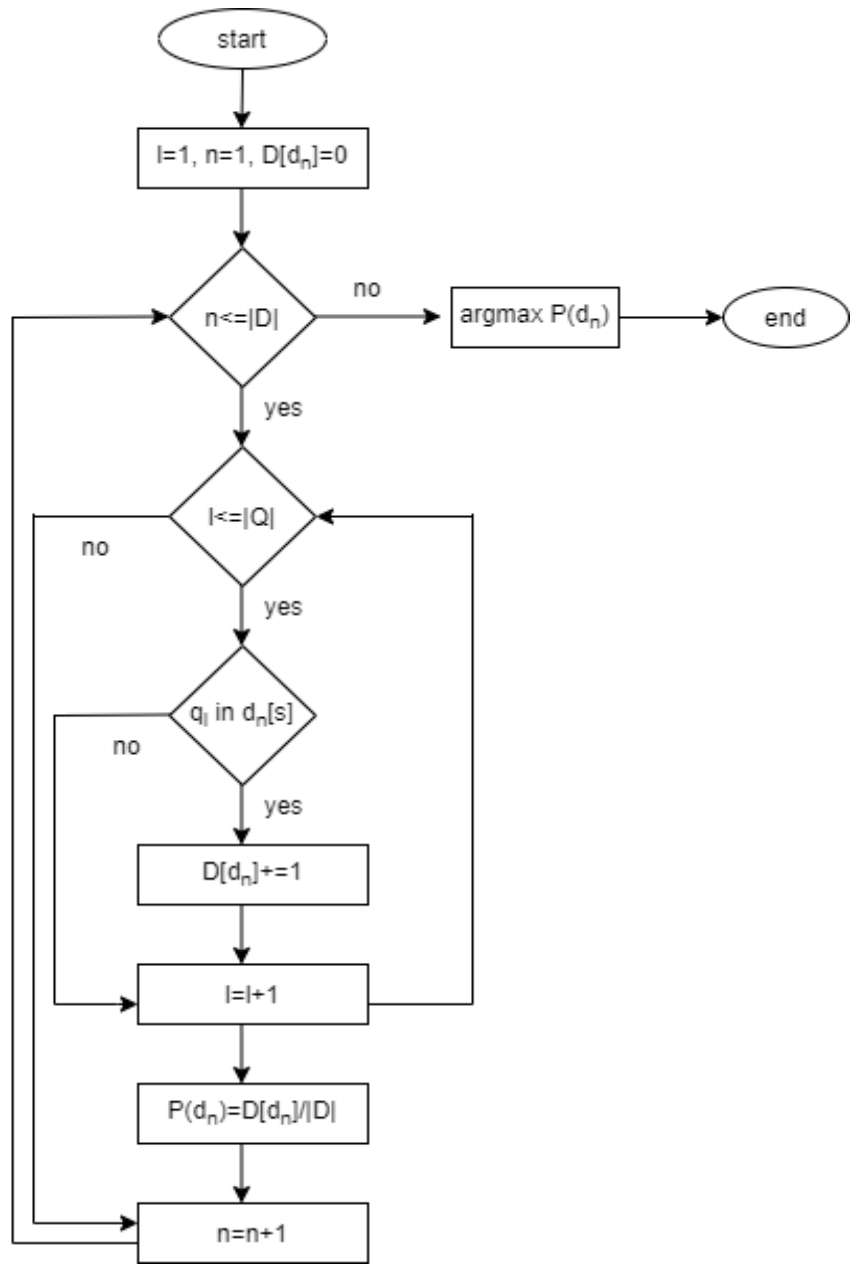

*Figure 2: Selection of disease based symptoms flowchart*

The above flowchart shows how a disease is suggested based on the second type of input, inserting set of symptoms. |D| represents the number of disease in the database. When the HEW feed queries to MatES, the system matches the queries, Q, with the symptoms, S, of the diseases. $d_n[s]$ represents the set of symptoms of $d_n$. The dictionary $D[d_n]$ stores the incidence of each query in list of symptoms of disease n, $d_n$. Finally, it computes the probability of each disease and suggests the diseases with highest probability. The output of this algorithm will be list of disease in descending order and it will be an input to the first algorithm presented in the form of pseudocode.

4. Implementation

MatES is an N-tier architecture. It consists the client layer, the application logic layer, the inference engine layer and the database layer. All components of the system are developed using HTML and PHP. The inference engine is developed using PHLIPS, PHP version of CLIPS. The web server handles requests from web clients and generate response through HTML document.

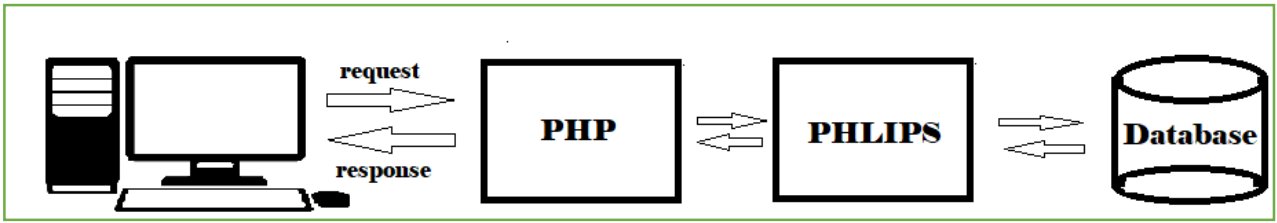

*Figure 3: N-Tier Architecture of MatES*

MatES represents the above two input scenarios and presents two main user interfaces. The first user interface shows list of diseases. The HEWs can select a number of diseases and consult the system. The inference engine fires the care and treatment based on the selected disease type. Figure 3 shows when the input of MatES is disease with its care and treatment, and the consequence if it is not treated.

The second user interface shows list of symptoms. This component of MatES is developed to assist the HEWs if the disease is not known but can be described by its symptoms. Figure 4 shows the second user interface. The first task of the expert system is mapping the symptoms with the list of diseases in the knowledge-base. And then compute the probability of each diseases and present the suggestion in descending order. The first disease is the first suggestion of MatES and continues up to the last suggestion. For each identified diseases MatES provides care and treatment, and consequences if it is not treated.

*Figure 4: MatES with disease as input*

| ☑ Cough | ☑ Fever | ☑ Weight Loss | ☐ Night Sweat |
| ☐ Hemoptysis | ☐ Malaise | ☐ Headache | ☑ Myalgia |
| ☐ Vomiting | ☐ Nausea | ☐ Jaundice | ☑ Fatigue |
| ☐ Shakiness | ☐ Feel Weak | ☐ Vaginal Discharge | ☐ Pelvic Pain |

| ☐ Changes in Appetite | ☐ Concentration Problems | ☐ A low Mood |
| ☐ Genital Ulcers | ☐ Abdominal Pain | ☐ Concetrated Urine |
| ☐ A sad Mood | ☐ Changes in Sleep | ☐ Changes in Energy |
| ☐ Problems Thinking | ☐ Urine that Smells Bad | ☐ Urine looks Cloudy |
| ☐ Urine looks Reddish | ☐ Pressure in Lower Belly | ☐ An Urge to use Bathroom often |
| ☐ Loss of interest in fun Activities | ☐ Pain while Using Bathroom | ☐ Lack of Appropraite Weight Gain |
| ☐ Problems in Making Decisions | ☐ Feelings of Worhtlessness | ☐ Feelings of Shame |
| ☐ Feelings of Guilt | ☐ Thoughts that Life is not worth Living | ☐ Back Pain |

*Figure 5: User interface for symptoms selection*

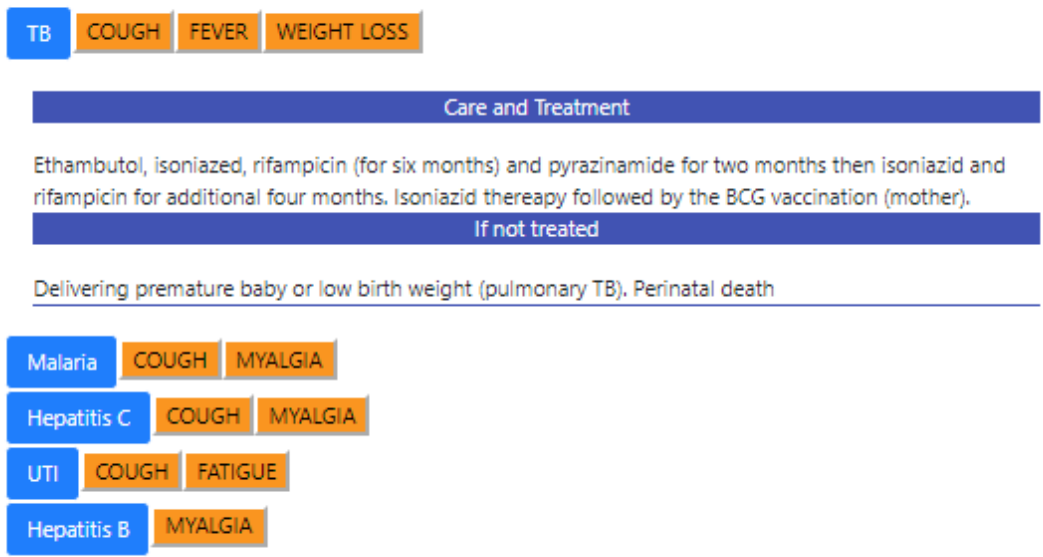

*Figure 6: User interface for disease suggestion*

MatES is tested by knowledge engineers as well as nurses and gynecologists. The test by knowledge engineers is conducted to ensure that every fact, rule and conditions are represented in the knowledge-base. The test conducted by nurses and gynecologists is to make sure that the system correctly suggests a disease, care and treatment, and consequences. The tests of the nurses and gynecologists are done on the two algorithms. The first system test is on algorithm 1. This test checks if the system generates a correct care and treatment as well as consequence if it is not treated. Since the system retrieves the care and treatment as well as consequence from the database that is uniquely identified by the disease, we got 100% correct results.

The accuracy of algorithm 2 depends on the number of selected symptoms. Diseases that contain at least one query from the set of symptoms is suggested. The disease with more number of symptoms selected will be at the top of the suggestions list as it shown in figure 6.

5. **Conclusion and Recommendation**

The web-based expert system for maternal care can serve as an assistant for HEWs and practical guideline for medical students. This web application uses forward chaining rule-based expert system to infer a new knowledge and reason. MatES is easy to use. Anyone who has basic knowledge of how computer or smart phone operates can use the system without any challenge. But for the sake of better understanding and usage of the results of the system, we advise the system to be used by health professionals. In this work, we only targeted 10 major diseases in Sub-Saharan Africa that create complications in maternal health. The system can be improved by considering more types of diseases and complications. In addition to that we used forward chaining rule-based expert system. Considering other type of expert systems may give better suggestions. Therefore, we recommend other to conduct experiments on the other types of expert systems.

# References

Ahmed A. Al-Hajji, F. M. (2019). An online expert system for psychiatric diagnosis. *International Journal of Artificial Intelligence and Applications (IJAIA)*, 59-76.

Berhan, Y. (2014). Review of maternal moratality in Ethiopia: A story of the past 30 years. *Ethiopian Journal of Health Science*, 3-14.

Bilal, N. K. (2011). Health extension workers in Ethiopia: improved access and coverage for the rural poor. *Yes Africa Can: Success Stiroes from a Dynamic Continent*, 433--443.

CDC. (2018, 10 23). *Pregnancy Complications | Maternal and Infant Health | CDC*. Retrieved from CDC: https://www.cdc.gov/reproductivehealth/maternalinfanthealth/pregnancy-complications.html

Daoliang Li, Z. F. (2002). Fish-Expert: a web-based expert system for fish disease diagnosis. *23*.

Desta, F. S. (2017). Identifying gaps in the practices of rural health extension workers in Ethiopia: a task analysis study. *BMC Health Services Research, 17*(839).

Fahad Shahbaz Khan, S. R. (2008). Dr. Wheat: A web-based expert system for diagnosis of diseases and pests in Pakistani wheat. London.

FAZEL, Z. M. (2010). A fuzzy rule-based expert system for diagnosing asthma. SCIENTIA IRANICA.

Fu Zetian, X. F. (2005). Pig-vet: a web-based expert system for pig disease diagnosis.

Gatot Supriyanto, A. G. (2019). Career Guidance Web-based expert system for vocational students. *14*(4).

Grensya Bella Vega Persulessy, N. S. (2019). Web-based expert system to detect stress on college students. *10*(1).

Keleş A, K. A. (2011). Expert system based on neuro-fuzzy rules for diagnosis breast cancer. *Expert systems with applications*, 5719--5726.

Persulessy, G. B., Pratama, N. S., Setiawan, N., & Sevani, N. (2019). Web-based expert system to detect stress on college students. *10*(1).

Ravisankar, H., Sivaraju, K., D, D. R., & Rao, S. (2018). Web based expert system for tobacco disease management. *Journal of Entomology and Zoology Studies*, 05-11.

Seto, E. a. (2012). Developing healthcare rule-based expert systems: case study of a heart failure telemonitoring system. *International journal of medical informatics*, 556--565.

WHO. (2019, 09 19). *Maternal moratlity*. Retrieved from WHO: https://www.who.int/news-room/fact-sheets/detail/maternal-mortality